\documentclass[runningheads]{llncs}
\usepackage[T1]{fontenc}
\usepackage{graphicx}
\usepackage{amsmath}
\usepackage{amssymb} 
\usepackage{booktabs}
\usepackage{amsmath}
\usepackage{algorithm}
\usepackage{amsmath}
\usepackage{algorithmic}
\usepackage{multirow}
\usepackage{subcaption}
\usepackage{array}

\begin{document}
\title{CoGS: Causality Constrained Counterfactual Explanations using goal-directed ASP\thanks{Authors supported by US NSF Grants IIS 1910131, US DoD, and industry grants. 
}}
%
%
\author{Sopam Dasgupta\inst{1} \and
Joaqu\'in Arias\inst{2} \and
Elmer Salazar\inst{3} \and
Gopal Gupta\inst{4}
}

\titlerunning{Causality Constrained Counterfactuals}
\authorrunning{S. Dasgupta et al.}
%
\institute{The University of Texas at Dallas, Richardson TX 75080, USA \and
Universidad Rey Juan Carlos, 28933 Móstoles, Madrid, Spain \and
The University of Texas at Dallas, Richardson TX 75080, USA \and
The University of Texas at Dallas, Richardson TX 75080, USA\\
}
\maketitle              
\begin{abstract}

Machine learning models are increasingly used in areas such as loan approvals and hiring, yet they often function as black boxes, obscuring their decision-making processes. Transparency is crucial, and individuals need explanations to understand decisions, especially for the ones not desired by the user. Ethical and legal considerations require informing individuals of changes in input attribute values (features) that could lead to a desired outcome for the user. Our work aims to generate counterfactual explanations by considering causal dependencies between features. We present the \textit{CoGS (Counterfactual Generation with s(CASP))} framework that utilizes the goal-directed Answer Set Programming system s(CASP) to generate counterfactuals from rule-based machine learning models, specifically the FOLD-SE algorithm. \textit{CoGS} computes realistic and causally consistent changes to attribute values taking causal dependencies between them into account. It finds a path from an undesired outcome to a desired one using counterfactuals. We present details of the \textit{CoGS} framework along with its evaluation. 

\vspace{-0.12in}
\end{abstract}
\section{Introduction}
\vspace{-0.1in}


Predictive models are widely used in automated decision-making processes, such as job-candidate filtering or loan approvals. These models often function as black boxes, making it difficult to understand their internal reasoning for decision making. The decisions made by these models can have significant consequences, leading individuals to seek satisfactory explanations, especially for an unfavorable decision. This desire for transparency is crucial, whether the decision is made by an automated system or humans. Explaining these decisions presents a significant challenge. Additionally, users want to know what changes they must make to flip an undesired (negative) decision into a desired (positive) one.



In this work we follow Wachter et al.'s \cite{wachter} approach where \textit{counterfactuals} are employed to explain the reasoning behind a prediction by a machine learning model. Counterfactuals help answer the following question: ``What changes should be made to input attributes or features to flip an undesired outcome to a desired one?" Counterfactuals also serve as a good candidate for explaining a prediction. Wachter et al. \cite{wachter} use statistical techniques, where they examine the proximity of points in the N-dimensional feature space, to find counterfactuals. We present the \textit{Counterfactual Generation with s(CASP) (CoGS)} framework, which generates counterfactual explanations from \textit{rule-based machine learning (RBML)} algorithms such as FOLD-SE \cite{foldse}. Our framework makes two advances compared to Wachter et al.'s work:  (i) It computes counterfactuals using RBML algorithms and ASP, rather than statistical techniques, (ii) It takes causal dependencies among features into account when computing these counterfactuals. Another novelty of the \textit{CoGS} framework is that it further leverages the FOLD-SE algorithm \cite{foldse} to automatically discover potential dependencies between features that are subsequently approved by a user. 


Our approach models various scenarios (or worlds): the current \textit{initial state} $i$ representing negative a outcome and the \textit{goal state} $g$ representing a positive outcome. A state is represented as a set of feature-value pairs. \textit{CoGS} finds a path from the \textit{initial state} $i$ to the \textit{goal state} $g$ by performing interventions (or transitions), where each intervention corresponds to changing a feature value while taking causal dependencies among features into account. These interventions ensure realistic and achievable changes that will take us from state $i$ to $g$. \textit{CoGS} relies on common-sense reasoning, implemented through answer set programming (ASP) \cite{gelfond-kahl}, specifically using the goal-directed s(CASP) ASP system \cite{scasp-iclp2018}. The problem of finding these interventions can be viewed as a planning problem \cite{gelfond-kahl}, except that unlike the planning problem, the moves (interventions) that take us from one state to another are not mutually independent.

\vspace{-0.2 in}
\section{Background}\label{sec_background}
\vspace{-0.1 in}
\subsubsection{Counterfactual Reasoning:}\label{sec_cf}
%
Explanations help us understand decisions and inform actions. Wachter et al. \cite{wachter} advocated using counterfactual explanations (CFE) to explain individual decisions, offering insights on how to achieve desired outcomes. For instance, a counterfactual explanation for a loan denial might state: If John had \textit{good} credit, his loan application would be approved. This involves imagining alternate (reasonably plausible) scenarios where the desired outcome is achievable.
%
%
For a binary classifier given by $f:X \rightarrow \{0,1\}$, we define a set of counterfactual explanations $\hat{x}$ for a factual input $x \in X$ as $\textit{CF}_{f}(x)=\{\hat{x} \in X | f(x) \neq f(\hat{x})\}$. 
This set includes all inputs $\hat{x}$ leading to different predictions than the original input $x$ under $f$.
We utilize the s(CASP) query-driven predicate ASP system \cite{scasp-iclp2018} for counterfactual reasoning, incorporating \textit{causal dependency} between features. s(CASP) computes \textit{dual rules} (section \ref{dual_rules}), allowing negated queries and constructing alternate worlds/states that lead to counterfactual explanations while considering \textit{causal dependencies}.

\smallskip\noindent\textbf{Causality Considerations:}
Causality relates to cause-effect relationship among predicates. $P$ is the cause of $Q$, if $(P \Rightarrow Q)$ $\wedge$ $(\neg P \Rightarrow \neg Q)$ \cite{SCM}. We say that $Q$ is causally dependent on $P$. 
%
%
For generating realistic counterfactuals, causal dependencies among features must be taken into account. 
For example, in a loan approval scenario, increasing the \textit{credit score} to be `high' while still being under increasing \textit{debt} obligations is unrealistic due to the causal dependencies between \textit{debt}, and \textit{credit score}. Ignoring these dependencies could lead to invalid counterfactuals that do not achieve the desired outcome. Therefore, realistic counterfactual explanations must model these causal relationships to account for downstream effects of feature changes.

\smallskip\noindent\textbf{ASP, s(CASP) and Commonsense Reasoning:}\label{dual_rules}
%
Answer Set Programming (ASP) is a paradigm for knowledge representation and reasoning \cite{cacm-asp,baral,gelfond-kahl}, widely used in automating commonsense reasoning. We employ ASP to encode feature knowledge, decision-making rules, and causal rules, enabling the automatic generation of counterfactual explanations using this symbolic knowledge.
%
\textbf{s(CASP)} is a goal-directed ASP system that executes answer set programs in a top-down manner without grounding \cite{scasp-iclp2018}. Its query-driven nature aids in commonsense and counterfactual reasoning,
utilizing proof trees for justification. 
To incorporate negation-as-failure, s(CASP) adopts \textit{program completion}, turning ``if'' rules into ``if and only if'' rules. For every rule in a s(CASP) program, its corresponding \textit{dual} rule(s) is computed. Details can be found elsewhere \cite{scasp-iclp2018}.

\smallskip\noindent\textbf{FOLD-SE:}
%
Wang and Gupta \cite{foldse} developed FOLD-SE, an efficient, explainable \textit{rule-based machine learning (RBML)} algorithm for classification tasks. 
FOLD-SE generates default rules—--a stratified normal logic program—--as an \textit{explainable} model from the given input dataset. Both numerical and categorical features are allowed. The generated rules symbolically represent the machine learning model that will predict a label, given a data record. FOLD-SE can also be used for learning rules capturing causal dependencies among features in a dataset. FOLD-SE maintains scalability and explainability, as it learns a relatively small number of learned rules and literals regardless of dataset size, while retaining good classification accuracy compared to state-of-the-art machine learning methods.
%
In \textit{CoGS}, FOLD-SE is used to learn \textit{causal rules} that accurately model feature dependencies and that are subsequently used to generate realistic counterfactual explanations.

\smallskip\noindent\textbf{The Planning Problem:}
%
Planning involves finding a sequence of transitions from an initial state to a goal state while adhering to constraints. In ASP, this problem is encoded in a logic program with rules defining transitions and constraints restricting the allowed transitions \cite{gelfond-kahl}. Solutions are represented as a series of transitions through intermediate states. Each state is represented as a set of facts or logical predicates. Solving the planning problem involves searching for a path of transitions that meets the goal conditions within the constraints.
\textit{CoGS} can be thought of as a framework to find a plan---a series of interventions that change feature values---that will take us from the initial state to the final goal state. However, unlike the planning domain, the interventions (moves) are not independent of each other due to causal dependencies among features.

\vspace{-0.15 in}
\section{Overview}\label{sec_overview}
\vspace{-0.07 in}

\subsection{The Problem}
\vspace{-0.05 in}

When an individual (represented as a set of features) receives an undesired negative decision (loan denial), they can seek necessary changes to flip it to a positive outcome. \textit{CoGS} automatically identifies these changes.
%
%
For example, if John is denied a loan (\textit{initial state $i$}), \textit{CoGS} models the (positive) scenarios (\textit{goal set $G$}) where he obtains the loan. Obviously, the (negative) decision in the \textit{initial state $i$} should not apply to any scenario in the \textit{goal set $G$}. The query goal `{\tt ?- reject\_loan(john)}' should be {\tt True} in the initial state \(i\) and {\tt False} for all goals in the \textit{goal set $G$}. The problem is to find a series of intervensions, namely, changes to feature values, that will take us from $i$ to $g \in G$.
\vspace{-0.15 in}
\subsection{Solution: \textit{CoGS} Approach}\label{sec_CoGS_approach}
\vspace{-0.05 in}

Inspired by the planning problem, the \textit{CoGS} approach casts the solution as traversing from an \textit{initial state} to a \textit{goal state}, represented as feature-value pairs (e.g., credit score: 600; age: 24). There can be multiple goal states that represents the positive outcome (set of goal states $G$). The objective is to turn a negative decision (\textit{initial state i}) into a positive one (\textit{goal state g}) through necessary changes to feature values, so that the query goal `{\tt ?- not reject\_loan(john)}' will succeed for $g \in G$. As mentioned earlier, the \textit{CoGS} framework involves solving a variant of the planning problem due to the dependence between features.


\textit{CoGS} models two scenarios: 1) the negative outcome world (e.g., loan denial, initial state \(i\)), and 2) the positive outcome world (e.g., loan approval, \textit{goal state $g$}) achieved through specific interventions. The initial state \(i\) and \textit{goal state $g$} are defined by specific attribute values (e.g., loan approval requires a credit score >=600). \textit{CoGS} symbolically computes the necessary interventions to find a path to the \textit{goal state $g$}, representing a flipped decision.


Given an instance where the decision query (e.g., `{\tt ?- reject\_loan/1}') succeeds (negative outcome), \textit{CoGS} finds the state where this query fails (i.e., the decision query `{\tt ?- not reject\_loan/1}' succeeds), which then constitutes the \textit{goal state $g$}. 
In terms of ASP, the problem of finding interventions can be cast as follows: given a possible world where a query succeeds, compute changes to the feature values (while taking their causal dependencies into account) that will reach a possible world where negation of the query will succeed. Each of the intermediate possible worlds we traverse must be viable worlds with respect to the rules. 
%
We use the s(CASP) query-driven predicate ASP system \cite{scasp-iclp2018} for this purpose. s(CASP) automatically generates dual rules, allowing us to execute negated queries (such as `{\tt ?- not reject\_loan/1}') constructively.


\textit{CoGS} employs two kinds of actions: 1) Direct Actions: directly changing a feature value, and 2) Causal Actions: changing other features to cause the target feature to change, utilizing the causal dependencies between features. These actions guide the individual from the \textit{initial state $i$} to the \textit{goal state $g$} through (consistent) intermediate states, suggesting realistic and achievable changes.
Unlike \textit{CoGS}, Wachter et al.'s approach \cite{wachter} can output non-viable solutions.

\vspace{-0.15 in}
\subsubsection{Example 1- Using direct actions to reach the counterfactual state:} \label{eg_1}
Consider a loan application scenario. There are 4 feature-domain pairs: 1) Age: \{1 year,..., 99 years\}, 2) Debt: \{$\$1$, ..., $\$1000000$\}, 3) Bank Balance: \{$\$0$, ..., $\$1\ billion$\} and 4) Credit Score: \{$300\ points, ..., 850\ points$\}.

\begin{figure}[htp]
    \centering    \includegraphics[width=12cm]{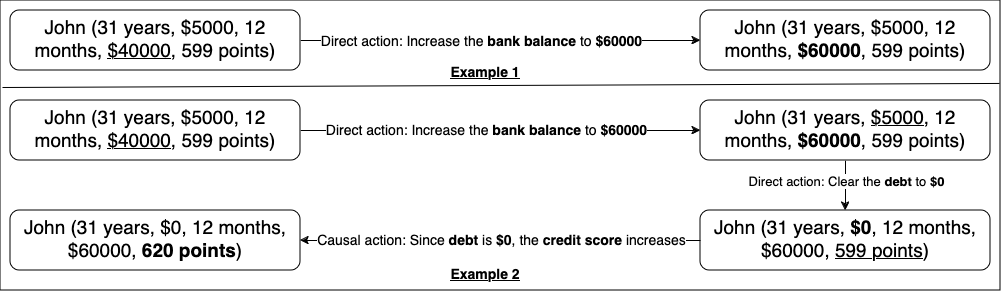}
    \caption{\textbf{Top:} Example 1 shows how John goes from being rejected for a loan to having his loan approved. Here the bank only considers the \textit{bank balance} for loan approval. John does a direct action to increase his \textit{bank balance} to $\$60000$. \textbf{Bottom:} Example 2 shows how John goes from being rejected for a loan to having his loan approved. Here the bank considers both \textit{bank balance} as well as \textit{credit score} for loan approval. While the \textit{bank balance} is directly altered by John, altering the \textit{credit score} requires John to directly alter his \textit{debt} obligations first. After clearing his \textit{debt}, the causal effect of having $\$0$ \textit{debt} increases John's \textit{credit score} to $620\ point$. This is the causal action\vspace{-0.2in}}
    \label{fig_example}
\end{figure}

John (31 years, $\$5000$, $\$40000$, $599\ points$) applies for a loan. The bank's rule is to deny loans to individuals with a \textit{bank balance} of \textbf{less than $\$60000$}. Hence, John is denied a loan (negative outcome). To get his loan approved (positive outcome), \textit{CoGS} suggests the following: \textbf{Initial state}: John (31 years, $\$5000$, 12 months, $ \$40000$, $599\ points$) is denied a loan. \textbf{Goal state}: John (31 years, $\$5000$, 12 months, $\$60000$, $599\ points$) is approved for a loan. \textbf{Intervention}: Suggest to John to change his \textit{bank balance} to $\$60000$.
%
%
As shown in Fig. \ref{fig_example} (top), the direct action of increasing the \textit{bank balance} to $\$60000$ flips the decision. John becomes eligible for the loan, bypassing bank’s rejection criteria.

\vspace{-0.2 in}

\smallskip\noindent\textbf{Example 2- Highlighting the utility of Causal Actions:}
This example demonstrates the advantages of incorporating causal rules. Consider a scenario where the bank has two rejection rules: 1) Deny loans to individuals with a \textit{bank balance} of \textbf{less than $\$60000$}, and 2) Deny loans to individuals with a \textit{credit score} below $600$. John (31 years, $\$5000$,  $ \$40000$, $599\ points$) is denied a loan (negative outcome) but wishes to get it approved (positive outcome).
%
%
Without the knowledge of causal dependencies, the solution will be the following: \textbf{Initial state}: John (31 years, $\$5000$, 12 months, $\$40000$, $599\ points$) is denied a loan. \textbf{Goal state}: John (31 years, $\$5000$, 12 months, $\$60000$, $620\ points$) is approved for a loan. \textbf{Interventions}: 1) Change the \textit{bank balance} to $\$60000$, and 2) the \textit{credit score} to $620\ points$. However, \textit{credit score} cannot be changed directly.



To realistically increase the credit score, the bank's guidelines suggest \textbf{1) having no debt}. This leads to a causal dependency between \textit{debt} and \textit{credit score}. Incorporating this, \textit{CoGS} provides: \textbf{Initial state}: John (31 years, $\$5000$, 12 months, $ \$40000$, $599\ points$) is denied a loan. \textbf{Goal state}: John (31 years, $\$0$, 12 months, $\$60000$, $620\ points$) is approved for a loan. \textbf{Interventions}: 1) John changes his \textit{bank balance} to $\$60000$, and 2) \textit{reduces his debt} to \$0 to increase his \textit{credit score}. 



As shown in Figure \ref{fig_example} (bottom), by clearing the \textit{debt} (direct action), John's \textit{credit score} increases (causal effect), making him eligible for the loan. Intermediate states, such as John with $\$5000$ in \textit{debt} and John with $\$0$ in \textit{debt}, are part of the path to the goal state. This example illustrates how using causal dependencies between features allows realistic achievement of desired outcomes.



Hence, we demonstrate how utilizing causal dependencies between features allows us to realistically achieve desired outcomes by making appropriate changes. The challenge now is to accomplish this automatically, i.e.,: (i) identify causal dependencies automatically: we use a rule-based machine learning algorithm (FOLD-SE) for this purpose. (ii) Compute the sequence of necessary interventions automatically: in particular, we want to avoid repeating states, a known issue in the planning domain.
Our \textit{CoGS} approach addresses these challenges. It generates the path from the \textit{initial state i} to the counterfactual \textit{goal state g} using the \textit{find\_path} algorithm (Algorithm \ref{alg_path}), explained in Section \ref{sec_alg_path}.

\vspace{-0.13 in}

\section{Methodology}\label{sec_methodology}
\vspace{-0.07 in}

We next outline the methodology used by \textit{CoGS} to generate paths from the initial state (\textit{negative outcome}) to the goal state (\textit{positive outcome}). Unlike traditional planning problems where actions are typically independent, our approach involves \textit{interdependent} actions governed by causal rules $C$. This ensures that the effect of one action can influence subsequent actions, making interventions realistic and causally consistent. 
%
%
Note that the \textit{CoGS} framework uses the FOLD-SE \textit{RBML} algorithm \cite{foldse} to automatically compute causal dependency rules. These rules have to be either verified by a human, or commonsense knowledge must be used to verify them automatically. This is important, as \textit{RBML} algorithms can identify a correlation as a causal dependency. \textit{CoGS} uses the former approach. The latter is subject of research. We next define specific terms.




\begin{definition}[\textbf{State Space (S)}]\label{defn:S}
$S$ represents all combinations of feature values. For domains \( D_1,..., D_n \) of the features \( F_1,..., F_n\), $S$ is a set of possible states $s$, where each state is defined as a tuple of feature values \( V_1, ..., V_n \)

\vspace{-0.2in}
\begin{equation}
s\in S\ where\  
S = \{ (V_1,V_2,...,V_n) | V_i \in D_i,\ for\ each\ i\ in\ 1,...,n \} \label{defn:1}
\end{equation}

\noindent For example state $s$ can be : $s = (31\ years,\ \$5000,\ \$40000,\ 599\ points)$.
\end{definition}

\begin{definition}[\textbf{Causally Consistent State Space ($S_C$)}]\label{defn:S_C}
$S_C$ is a subset of $S$ where all causal rules are satisfied. $C$ represents a set of causal rules over the features within a state space $S$. Then, $\theta_C: P(S) \rightarrow P(S)$ (where $P(S)$ is the power set of S) is a function that defines the subset of a given state sub-space $S'\subseteq S$ that satisfy all causal rules in C.

\vspace{-0.1 in}
\begin{equation}
\theta_C(S') = \{ s \in S' \mid s\ satisfies\ all\ causal\ rules\ in\ C\} \label{defn:2_1}
\end{equation}
\begin{equation}
S_C = \theta_C(S) \label{defn:2_2}
\end{equation}



\noindent 
E.g., if a causal rule states that if $debt$ is $0$, the credit score should be above 599, then instance $s_1 = (31\ years,\ \$0,\ \$40000,\ 620\ points)$ is causally consistent, and instance $s_2 = (31\ years,\ \$0,\ \$40000,\ 400\ points)$ is causally inconsistent. 

\end{definition}

\noindent 
In a traditional planning problem, allowed actions in a given state are independent, i.e., the result of one action does not influence another. In \textit{CoGS}, causal actions are interdependent, governed by $C$. 

\vspace{-0.1 in}

\begin{definition}[\textbf{Decision Consistent State Space ($S_Q$)}]\label{defn:S_Q}
%
$S_Q$ is a subset of $S_C$ where all decision rules are satisfied. $Q$ represents a set of rules that compute some external decision for a given state. $\theta_Q: P(S) \rightarrow P(S)$ is a function that defines the subset of the causally consistent state space $S'\subseteq S_C$ that is also consistent with decision rules in $Q$:

\vspace{-0.1 in}
\begin{equation}
\theta_Q(S') = \{ s \in S' \mid s\ satisfies\ any\ decision\ rule\ in\ Q\} \label{defn:3_1}
\end{equation}
Given $S_C$ and $\theta_Q$, we define the decision consistent state space $S_Q$ as 
\vspace{-0.1 in}
\begin{equation}
S_Q = \theta_Q(S_C) = \theta_Q(\theta_C(S)) \label{defn:3_2}
\end{equation}

\noindent For example, an individual John whose loan has been rejected: \textit{s = ($31$ years, $\$0$, $\$40000$, $620$ points)}, where\ $s\in S_Q$.

\end{definition}

\vspace{-0.1 in}
\begin{definition}[\textbf{Initial State ($i$)}]\label{defn:I}
$i$ is the starting point with an undesired outcome. Initial state $i$ is an element of the \textit{causally consistent state space} $S_C$
\vspace{-0.1 in}
\begin{equation}
i \in S_C \label{defn:4}
\end{equation}
\vspace{-0.15in}

%
\noindent For example, $i = (31\ years,\ \$0,\ \$40000,\ 620\ points)$
\end{definition}

\vspace{-0.1in}
\begin{definition}[\textbf{Actions}]\label{defn:action}
The set of actions $A$ includes all possible interventions (actions) that can transition a state from one to another within the state space. Each action $a\in A$ is defined as a function that maps $s$ to a new state $s'$. 
\begin{equation}
    a: S\rightarrow S\ \mid where\ a\in A
\end{equation}

Actions are divided into: 1) Direct Actions: Directly change the value of a single feature of a state $s$, e.g.,  Increase bank balance from $\$40000$ to $\$60000$. 2) Causal Actions: Change the value of a target feature by altering related features, based on causal dependencies. It results in a causally consistent state with respect to C, e.g., reduce debt to increase the credit score. 
\end{definition}

\vspace{-0.1in}
\begin{definition}[\textbf{Transition Function}]\label{defn:delta}
A transition function $\delta: S_C \times A \rightarrow S_C$ maps a causally consistent state to the set of allowable causally consistent states that can be reached in a single step, and is defined as: 


  \[\delta(s,a) = \left\{\begin{array}{l}
     a(s) \textit{ if } a(s) \in S_C\\
     \delta(a(s),a') \textit{ with } a\in A, a'\in A \textit{, otherwise}
  \end{array}\right.\]

\noindent $\delta$ models a function that repeatedly takes actions until a causally consistent state is reached. In \textbf{example 1}, $\delta$ suggests  changing the \textit{bank balance} from $\$40000$ to $ \$60000$: 
$\delta(31\ years,\$5000,\$40000,599) = $ $(31\ years,\$5000,\$60000,599)$


\end{definition}


\begin{definition}[\textbf{Counterfactual Generation (CFG) Problem}]\label{problem_statement}
A counterfactual generation (CFG) problem is a 4-tuple $(S_C,S_Q,I,\delta)$ where $S_C$ is causally consistent state space, $S_Q$ is the decision consistent state space, $I\in S_C$ is the initial state , and $\delta$ is a transition function.
\end{definition}

\begin{definition}[\textbf{Goal Set}]\label{defn:G}
The goal set $G$ is the set of desired outcomes that do not satisfy the decision rules $Q$. For counterfactual $(S_C,S_Q,I,\delta)$, $G \subseteq S_C$:
\begin{equation}
G = \{ s \in S_C| s\not\in S_Q\} \label{defn:5}
\end{equation}
$G$ includes all states in $S_C$ that do not satisfy $S_Q$. For \textbf{example 1}, an example goal state $g\in G$ is $g = (31\ years,\ \$5000,\ \$60000,\ 599\ points).$

\end{definition}

\begin{definition}[\textbf{Solution Path}]\label{solution_path}
A solution to the problem $(S_C,S_Q,I,\delta)$ with Goal set $G$ is a path:
\vspace{-0.1 in}
\begin{equation} \label{defn:8}
\begin{split}
s_0,s_1,...s_m\ where\ s_j\in S_C\ for\ all\ j\ \in\{0,...,m\}\ such\ that \\
s_0,...,s_{m-1} \not\in G;\ s_{i+1}\in \delta(s_i)\ for\ i\ \in\{0,...,m-1\};\ s_0 = I; s_m\in G 
\end{split}
\end{equation}

\vspace{-0.05 in}

For \textbf{example 1}, individuals with less than $\$60000$ in their account are ineligible for a loan, thus the state of an ineligible individual $s\in S_Q$ might be $s = (31\ years,\$5000,\$40000,599\ points)$. The goal set has only one goal state $g\in G$ given by $s = (31\ years,\$5000,\$60000,599\ points)$. The path from $s$ to $g$ is \{\textit{(31\ years,\$5000,\$40000,599\ points)}$\rightarrow Direct\rightarrow $\textit{(31\ years,\$5000,\$60000,599\ points)}$\}$. Here, the path has only 2 states as only changing the \textit{bank balance} to be $\$60000$ is needed to reach the goal state.
\end{definition}

\vspace{-0.25 in}
\subsection{Algorithm}
\vspace{-0.05 in}

We next describe our algorithm to find the goal states and compute the solution paths. The algorithm makes use of the following functions: (i) \textbf{not\_member}: checks if an element is: \textit{a}) \textbf{not} a member of a list, and \textit{b}) Given a list of tuples, \textbf{not} a member of any tuple in the list. (ii) \textbf{drop\_inconsistent}: given a list of states [$s_0,...,s_k$] and a set of Causal rules $C$, it drops all the inconsistent states resulting in a list of consistent states with respect to $C$. (iii) \textbf{get\_last}: returns the last member of a list. (iv) \textbf{pop}: returns the last member of a list. (v) \textbf{is\_counterfactual}: returns {\tt True} if the input state is a causally consistent counterfactual solution (see supplement for details \cite{ref_supplement}).

\vspace{-0.25in}

\begin{algorithm}[h!]
\caption{\textbf{find\_path}: Obtain a path to the counterfactual state}
\label{alg_path}
\begin{algorithmic}[1]
    \REQUIRE Initial State $(I)$, States $S$, Causal Rules $C$, Decision Rules $Q$, \textit{is\_counterfactual} (Algorithm \ref{alg_counterfactual}), \textit{delta} (Algorithm \ref{alg_intervene}), Actions $a\in A$:

    \STATE\label{PS21} Create an empty list \textit{visited\_states} that tracks the list of states traversed (so that we avoid revisiting them).
    \STATE Append ($i$, [~]) to \textit{visited\_states} 
    \WHILE{$is\_counterfactual(get\_last(visited\_states),C,Q)\ is\ FALSE$ }
    \STATE Set $visited\_states=intervene(visited\_states,C,A) $
    \ENDWHILE
    \STATE \textit{candidate\_path = drop\_inconsistent(visited\_states)}
    \STATE Return \textit{candidate\_path}

\end{algorithmic}
\end{algorithm}
\vspace{-0.2 in}

\smallskip\noindent\textbf{Find Path:}\label{sec_alg_path}
%
Function `\textit{\textbf{find\_path}}' implements the Solution Path $P$ of Definition \ref{solution_path}. Its purpose is to find a path to the counterfactual state. Algorithm \ref{alg_path} provides the pseudo-code for `\textit{\textbf{find\_path}}', which takes as input an Initial State $i$, a set of Causal Rules $C$, decision rules $Q$, and actions $A$. It returns a path to the counterfactual state/goal state $g\in G$ for the given $i$ as a list `\textit{visited\_states}'. Unrealistic states are removed from `\textit{visited\_states}' to obtain a `\textit{candidate\_path}'.

Initially, $s=i$. The function checks if the current state $s$ is a counterfactual. If $s$ is already a counterfactual, `\textit{\textbf{find\_path}}' returns a list containing $s$. If not, the algorithm moves from $s=i$ to a new causally consistent state $s'$ using the `\textit{\textbf{intervene}}' function, updating `\textit{visited\_states}' with $s'$. It then checks if $s'$ is a counterfactual using `\textbf{\textit{is\_counterfactual}}'. If {\tt True}, the algorithm drops all inconsistent states from `\textit{visited\_states}' and returns the `\textit{candidate\_path}' as the path from $i$ to $s'$. If not, it updates `$current\_state$' to $s'$ and repeats until reaching a counterfactual/goal state $g$. The algorithm ends when `\textbf{\textit{is\_counterfactual}}' is satisfied, i.e., $s'=g\mid\ where\ g \in G$.

\vspace{-0.17 in}
\begin{equation}
\begin{split}
   s'\in G \mid\ by\ definition \\
   s_0,...,s_k,s' \mid s_0,...,s_k: \not\in G
\end{split}
\end{equation}

\vspace{-0.15 in}

\smallskip\noindent\textbf{Intervene:}
Function `\textit{\textbf{intervene}}' implements the transition function $\delta$ from Definition \ref{defn:delta}. It is called by `\textit{\textbf{find\_path}}' in line 4 of Algorithm \ref{alg_path}. The primary purpose of `\textit{\textbf{intervene}}' is to transition from the current state to the next state, ensuring actions are not repeated and states are not revisited. Its pseudo-code as well as a detailed exploration is available in the supplement.

\vspace{-0.3 in}
\begin{algorithm}[h!]
\caption{\textbf{make\_consistent}: reaches a consistent state}
\label{alg_inner_delta}
\begin{algorithmic}[1]
    \REQUIRE State $s$, Causal \textit{rules} $C$, List \textit{visited\_states} , \textit{actions\_taken}, Actions $a\in A$:
    \WHILE{$s$ does not satisfy all rules in $C$}
    \STATE Try to select a causal action $a$ ensuring \textit{not\_member(a(s),visited\_states)} and \textit{not\_member(a,actions\_taken)} are $TRUE$
    \IF{ causal action $a$ exists}
    \STATE Set \textit{(s,actions\_taken),visited\_states=update(s,visited\_states,actions\_taken,a)}
    \ELSE
        \STATE Try to select a direct action $a$ ensuring \textit{not\_member(a(s),visited\_states)} and \textit{not\_member(a,actions\_taken)} are $TRUE$
        \IF{ direct action $a$ exists}
        \STATE Set \textit{(s, actions\_taken), visited\_states=update(s,visited\_states,actions\_taken,a)}
        \ELSE
        \STATE //Backtracking 
        \IF {\textit{visited\_states} is empty } 
        \STATE \textit{EXIT with Failure}
        \ENDIF
        \STATE Set $(s, actions\_taken)$ = \textit{pop(visited\_states)}
        \ENDIF
    \ENDIF
    \ENDWHILE
    \STATE Return $(s,actions\_taken),visited\_states$ .

\end{algorithmic}
\end{algorithm}
\vspace{-0.2 in}

\vspace{-0.1 in}
\noindent\textbf{Make Consistent:}
The pseudo-code for `\textit{\textbf{make\_consistent}}' is specified in Algorithm \ref{alg_inner_delta}. It takes as arguments a current State $s$, a list \textit{actions\_taken}, a list \textit{visited\_states}, a set of Causal Rules $C$ and a set of actions $A$. Called by `\textit{\textbf{intervene}}' in line 12 of Algorithm \ref{alg_intervene}, `\textit{\textbf{make\_consistent}}' transitions from the current state to a new, causally consistent state.

\vspace{-0.19 in}
\subsubsection{Update}\label{sec_update}

Function `\textit{\textbf{update}}' tracks the list of actions taken and states visited to avoid repeating actions and revisiting states. Its pseudo-code is provided and explored in detail in the supplement.



\smallskip\noindent\textbf{Discussion:} A few points should be highlighted: \textbf{(i)} 
%
Certain feature values may be immutable or restricted, such as \textit{age} cannot decrease or \textit{credit score} cannot be directly altered. To respect these restrictions, we introduce \textit{plausibility constraints}. These constraints apply to the actions in our algorithms, ensuring realistic changes to the features. Since they do not add new states but restrict reachable states, they are represented through the set of available actions in Algorithms \ref{alg_path},  \ref{alg_intervene},  \ref{alg_inner_delta},  \ref{alg_update}.
\textbf{(ii)}
Similarly, \textit{CoGS} has the ability to specify the path length for candidate solutions. Starting with a minimal path length of 1, \textit{CoGS} can identify solutions requiring only a single change. If no solution exists, \textit{CoGS} can incrementally increase the path length until a solution is found. This ensures that the generated counterfactuals are both \textbf{minimal and causally consistent}, enhancing their practicality and interpretability. This is achieved via constraints on path length.

\vspace{-0.2 in}
\subsection{Soundness}
\vspace{-0.05 in}

    

    
\begin{definition}[CFG Implementation]\label{defn:impl_cfg}
When Algorithm \ref{alg_path} is executed with the inputs: Initial State $i$ (Definition \ref{defn:I}), States Space $S$ (Definition \ref{defn:S}), Set of Causal Rules $C$ (Definition \ref{defn:S_C}), Set of Decision Rules $Q$ (Definition \ref{defn:S_Q}), and Set of Actions $A$ (Definition \ref{defn:action}), a CFG problem $(S_C,S_Q,I,\delta)$ (Definition \ref{problem_statement}) with causally consistent state space $S_C$ (Definition \ref{defn:S_C}), Decision consistent state space $S_Q$ (Definition \ref{defn:S_Q}), Initial State $i$ (Definition \ref{defn:I}), the transition function $\delta$ (Definition \ref{defn:delta}) is constructed. 
    
\end{definition}


\begin{definition}[\textit{Candidate path}]\label{defn:candidate_path}
Given the counterfactual $(S_C,S_Q,I,\delta)$ constructed from a run of  algorithm \ref{alg_path}, the return value (candidate path) is the resultant list obtained from removing all elements containing states $s'\not\in S_C$.     
\end{definition}

\vspace{-0.1 in}
\noindent Definition \ref{defn:impl_cfg} maps the input of Algorithm \ref{alg_path} to a \textit{CFG problem} (Definition \ref{problem_statement}). \textit{Candidate path} maps the result of Algorithm \ref{alg_path} to a possible solution (Definition \ref{solution_path}) of the corresponding CGF problem. From Theorem \ref{theorem_soundness} \textit{(proof in supplement \cite{ref_supplement})}, the \textit{candidate path} (Definition \ref{defn:candidate_path}) is a solution to the corresponding \textit{CFG problem} implementation (Definition \ref{defn:impl_cfg}).\\

\vspace{-0.1 in}
\noindent \textit{Theorem 1} 
Soundness: Given a CFG $\mathbb{X}=(S_C,S_Q,I,\delta)$, constructed from a run of Algorithm \ref{alg_path} \& a corresponding candidate path $P$, $P$ is a solution path for $\mathbb{X}$. Proof: Given in the supplemental document \cite{ref_supplement}.

\vspace{-0.20 in}
\section{Experiments}\label{sec_experiments}
\vspace{-0.13 in}

We applied the \textit{CoGS} methodology to rules generated by the FOLD-SE algorithm (code on GitHub \cite{ref_supplement}). Our experiments use the German dataset \cite{german}, Adult dataset \cite{adult}, and  the Car Evaluation dataset \cite{car}. These are popular datasets found in the UCI Machine Learning repository \cite{ref_UCI}. The German dataset contains demographic data with labels for credit risk (`\textit{good}' or `\textit{bad}'), with records with the label `\textit{good}' greatly outnumbering those labeled `\textit{bad}'. The Adult dataset includes demographic information with labels indicating income (‘$=<\$50k/year$’ or ‘$>\$50k/year$’). The Car Evaluation dataset provides information on acceptability of a used car being purchased. We relabelled the Car Evaluation dataset to \textit{`acceptable'} and \textit{`unacceptable'} in order to generate the counterfactuals.

For the (imbalanced) German dataset, the learned FOLD-SE rules determine `\textit{good}' credit rating, with the undesired outcome being a `\textit{good}' rating, since the aim is to identify criteria making someone a credit risk (`\textit{bad}' rating). Additionally, causal rules are also learnt using FOLD-SE and verified (for example, if feature \textbf{`Job'} has value  \textit{`unemployed'}, then feature \textbf{`Present employment since'} should have the value  \textit{`unemployed/unskilled-non-resident'}). We learn the rules to verify these assumptions on cause-effect dependencies. 

By using these rules that identify individuals with a `\textit{good}' rating, we found a path to the counterfactuals thereby depicting steps to fall from a \textit{`good'} to a \textit{`bad'} rating in Table \ref{tbl_exp}. Similarly, we learn the causal rules as well as the rules for the undesired outcome for the \textit{Adult} dataset (undesired outcome: ‘$=<\$50k/year$’). For the \textit{Car Evaluation} dataset (undesired outcome: \textit{`unacceptable'}), we only learn the rules for the undesired outcome as there are no causal dependencies (FOLD-SE did not generate any either). 
Table \ref{tbl_exp} shows a path to the counterfactual goal state for a specific instance for each of these datasets. Note that the execution time for finding the counterfactuals is also reported. While we have only shown specific paths in Table \ref{tbl_exp}, our \textit{CoGS} methodology can generate all possible paths from an original instance to a counterfactual. 
Note that each path may represent a set of counterfactuals. This is because numerical features may range over an interval. Thus, \textit{CoGS} generates 240 sets of counterfactuals for the the German dataset, 112 for the  Adult dataset, and 78 for the Car Evaluation dataset (See Table \ref{tbl_counterfactual} in the supplement \cite{ref_supplement}).

\begin{table*}[t]
\centering
\fontsize{4}{6}\selectfont
\setlength{\tabcolsep}{2.5pt}

\begin{subtable}{\textwidth}
\centering
\begin{tabular}{@{}p{1.7cm} p{2cm} p{2.5cm} p{1cm} p{2.5cm} | p{1.2cm}@{}}
\toprule
Dataset & Features & Initial State & Action & Goal State & Time (ms) \\ 
\midrule
\multirow{7}{*}{\textit{german}} & Checking account status & $\geq\ 200$ & N/A & $\geq\ 200$ & \multirow{7}{*}{3236} \\ 
\cmidrule(lr){2-5}
 & Credit history & no credits taken/all credits paid back duly & N/A & no credits taken/all credits paid back duly & \\ 
\cmidrule(lr){2-5}
 & \textbf{Property} & real estate & \textbf{Direct} & car or other & \\ 
\cmidrule(lr){2-5}
 & Duration months & 7 & N/A & 7 & \\ 
\cmidrule(lr){2-5}
 & Credit amount & 500 & N/A & 500 & \\ 
\cmidrule(lr){2-5}
 & Job & unemployed & N/A & unemployed & \\  
\cmidrule(lr){2-5}
 & Present Employment Since & unemployed/unskilled-non-resident & N/A & unemployed/unskilled-non-resident & \\ 
\midrule
\multirow{4}{*}{\textit{car evaluation}} & persons & 4 & N/A & 4 & \multirow{4}{*}{1221} \\ 
\cmidrule(lr){2-5}
 & \textbf{maint} & low & \textbf{Direct} & medium & \\ 
\cmidrule(lr){2-5}
 & buying & medium & N/A & medium & \\ 
\cmidrule(lr){2-5}
 & safety & medium & N/A & medium & \\ 
\bottomrule
\end{tabular}
\end{subtable}


\begin{subtable}{\textwidth}
\centering
\begin{tabular}{@{}p{0.8cm} p{1.9cm} p{1.6cm} p{0.7cm} p{1.6cm} p{0.8cm} p{2.2cm} | p{1.2cm}@{}}
\toprule
Dataset & Features & Initial State & Action & Intermediate & Action & Goal State & Time (ms) \\ 
\midrule
\multirow{6}{*}{\textit{adult}} & \textbf{Marital\_Status} & never\_married & N/A & never\_married & \textbf{Causal} & married\_civ\_spouse & \multirow{6}{*}{1126} \\
\cmidrule(lr){2-7}
 & Capital Gain & \$6000 & N/A & N/A & N/A & $> 6849$ and $\leq 99999$ & \\ 
\cmidrule(lr){2-7}
 & Education\_num & $7$ & N/A & N/A & N/A & $7$ & \\ 
\cmidrule(lr){2-7}
 & \textbf{Relationship} & unmarried & \textbf{Direct} & husband & N/A & husband & \\ 
\cmidrule(lr){2-7}
 & Sex & male & N/A & N/A & N/A & male & \\ 
\cmidrule(lr){2-7}
 & Age & 28 & N/A & N/A & N/A & 28 & \\ 
\bottomrule
\end{tabular}
\end{subtable}
\caption{Paths showing transitions to goal states alongside the time taken across different datasets: \textbf{1)} \textit{German:} The value of \textit{Property} changes from \textit{real estate} to \textit{car or other}. \textbf{2)} \textit{Car Evaluation:} The value of \textit{maint} goes from \textit{low} to \textit{medium}. \textbf{3)} \textit{Adult:} The value of \textit{Relationship} changes from \textit{unmarried} to \textit{husband}. This has a causal effect of altering \textit{Marital Status} to \textit{married\_civ\_spouse}.}
\label{tbl_exp}
\vspace{-0.4 in}
\end{table*}

\vspace{-0.17 in}
\section{Related Work and Conclusion}
\vspace{-0.09 in}
Various methods for generating counterfactual explanations in machine learning have been proposed. Wachter et al. \cite{wachter} aimed to provide transparency in automated decision-making by suggesting changes individuals could make to achieve desired outcomes. However, they ignored causal dependencies, resulting in unrealistic suggestions. Utsun et al. \cite{ref_2_ustun} introduced algorithmic recourse, offering actionable paths to desired outcomes but assuming feature independence, which is often unrealistic. \textit{CoGS} rectifies this by incorporating causal dependencies. Karimi et al. \cite{alt_karimi} focused on feature immutability and diverse counterfactuals, ensuring features like gender or age are not altered and maintain model-agnosticism. 
 However, this method also assumes feature independence, limiting realism.
Existing approaches include model-specific, optimization-based approaches  \cite{ref_mace_1,ref_mace_2}. White et al. \cite{ref_clear} showed how counterfactuals can enhance model performance and explanation accuracy . Karimi et al. \cite{ref_4_karimi_2} further emphasized incorporating causal rules in counterfactual generation for realistic and achievable interventions. 
However their method did not use the `if and only' property which is vital in incorporating the effects of causal dependence. Bertossi and Reyes \cite{ref_asp_cf}  rectified this by utilizing Answer Set Programming (ASP) but they relied on grounding, which can disconnect variables from their associations. In contrast, \textit{CoGS} does not require grounding as it leverages s(CASP) to generate counterfactual explanations, providing a clear path from undesired to desired outcomes. 



Eiter et al. \cite{ref_contrastive} proposed a framework for generating contrastive explanations in the context of ASP, focusing on why one particular outcome occurred instead of another. 
While contrastive explanations identify and compare the assumptions leading to different outcomes, \textit{CoGS} goes further by incorporating causal dependencies, ensuring that the generated counterfactuals are realistic and achievable. 

The main contribution of this paper is the Counterfactual Generation with s(CASP) (\textit{CoGS}) framework for automatically generating counterfactuals while taking causal dependencies into account to flip a negative outcome to a positive one. \textit{CoGS} has the ability to find minimal paths by iteratively adjusting the path length.
This ensures that explanations are both minimal and causally consistent. \textit{CoGS} is flexible, generating counterfactuals irrespective of the underlying rule-based machine learning (\textit{RBML}) algorithm. The causal dependencies can be learned from data using any \textit{RBML} algorithm, such as FOLD-SE. The goal-directed s(CASP) ASP system plays a crucial role, as it allows us to compute a possible world in which a query {\tt Q} fails by finding the world in which the query {\tt not Q} succeeds. \textit{CoGS} advances the state of the art by combining counterfactual reasoning, causal modeling, and ASP-based planning, offering a robust framework for realistic and actionable counterfactual explanations. Our experimental results show that counterfactuals can be computed for complex models in a reasonable amount of time. Future work will explore computing counterfactuals for image classification tasks, inspired by Padalkar et al \cite{parth_nesy}.

%
%
%
\vspace{-0.17in}
\bibliographystyle{splncs04}
\bibliography{bibliography}
\vspace{-0.07in}

%





\vfill \eject
\section{Supplementary Material}

\subsection{Methodology: Details}
\subsubsection{\textit{is\_counterfactual}: Checks for Goal State/ Counterfactual State}
\begin{algorithm}[h!]
\caption{\textbf{is\_counterfactual}: checks if a state is a counterfactual/goal state}
\label{alg_counterfactual}
\begin{algorithmic}[1]
    \REQUIRE State $s\in S$, Set of Causal \textit{rules} $C$, Set of Decision \textit{rules} $Q$
    \IF{$s$ satisfies \textbf{ALL} rules in $C$ \textbf{AND} $s$ satisfies \textbf{NO} rules in $Q$}
    \STATE Return $TRUE$.
    \ELSE    
    \STATE Return $FALSE$.
    \ENDIF
\end{algorithmic}
\end{algorithm}
The function \textit{is\_counterfactual} is our algorithmic implementation of checking if a state $s\in G$ from definition \ref{defn:G}.
In Algorithm \ref{alg_counterfactual}, we specify the pseudo-code for a function \textit{is\_counterfactual} which takes as arguments a state $s\in S$, a set of causal rules $C$, and a set of Decision rules $Q$. The function checks if a state $s\in S$ is a counterfactual/goal state. By definition \textit{is\_counterfactual} is $TRUE$ for state $s$ that is causally consistent with all $c\in C$ and \textbf{does not} agree with the any decision rules $q\in Q$.

\begin{equation}
   is\_counterfactual(s,C,Q)=TRUE \mid s\ agrees\ with\ C;\ s\ disagrees\ with\ Q; 
\end{equation}

\subsubsection{Intervene: Transition Function for moving from the current state to the new state}\label{sec_intervene}
\begin{algorithm}[h!]
\caption{\textbf{intervene}: reach a causally consistent state from a causally consistent current state}
\label{alg_intervene}
\begin{algorithmic}[1]
    \REQUIRE Causal \textit{rules} $C$, List \textit{visited\_states}, List \textit{actions\_taken}, Actions $a\in A$:
    \begin{itemize}
    
    \item Causal Action: $s$ gets altered to a causally
    consistent new state $s'=a(s)$. OR 
    \item Direct Action: new state $s'=a(s)$ is obtained by altering 1 feature value of $s$.     
    \end{itemize}

    \STATE Set $(s, actions\_taken)$ = \textit{pop(visited\_states)}
    \STATE Try to select an action $a\in A$ ensuring \textit{not\_member(a(s),visited\_states)} and \textit{not\_member(a,actions\_taken)} are $TRUE$
    \IF{ $a$ exists}
    \STATE Set $(s, actions\_taken),visited\_states=update(s,visited\_states,actions\_taken,a)$
    \ELSE
    \STATE //Backtracking 
    \IF {\textit{visited\_states} is empty }
    \STATE \textit{EXIT with Failure}
    \ENDIF
    \STATE Set $(s, actions\_taken)$ = \textit{pop(visited\_states)}
    \ENDIF
    \STATE Set $(s,actions\_taken), visited\_states=$\\ \hspace{0.2 in} $make\_consistent(s,actions\_taken, visited\_states,C,A)$ 
    \STATE Append $(s,actions\_taken)$ to \textit{visited\_states}
    \STATE Return \textit{visited\_states}.

\end{algorithmic}
\end{algorithm}
The function \textit{intervene} is our algorithmic implementation of the transition function $\delta$ from definition \ref{defn:delta}. In Algorithm \ref{alg_intervene}, we specify the pseudo-code for a function \textit{intervene}, which takes as arguments an Initial State $i$ that is causally consistent, a set of Causal Rules $C$, and a set of actions $A$. It is called by \textit{find\_path} in line 4 of algorithm \ref{alg_path}.

The function \textit{intervene} acts as a transition function that inputs a list \textit{visited\_states} containing the current state $s$ as the last element, and returns the new state $s'$ by appending $s'$ to \textit{visited\_states}. The new state $s'$ is what the current state $s$ traverses. Additionally, the function \textit{intervene} ensures that no states are revisited. In traversing from $s$ to $s'$, if there are a series of intermediate states that are \textbf{not} causally consistent, it is also included in \textit{visited\_states}, thereby depicting how to traverse from one causally consistent state to another.

\subsubsection{Update}
\begin{algorithm}[h!]
\caption{\textbf{update}: Updates the list \textit{actions\_taken} with the planned action. Then updates the current state.}
\label{alg_update}
\begin{algorithmic}[1]
    \REQUIRE State $s$, List \textit{visited\_states}, List \textit{actions\_taken}, Action $a \in A$:
    \begin{itemize}
        \item Causal Action: $s$ gets altered to a causally
        consistent new state $s'=a(s)$. OR 
        \item Direct Action: new state $s'=a(s)$ is obtained by altering 1 feature value of $s$.  
    \end{itemize}
    \STATE Append $a$ to \textit{actions\_taken}.
    \STATE Append $(s, actions\_taken)$ to \textit{visited\_states}.
    \STATE Set $s=a(s)$.
    \RETURN $(s, [~])$, \textit{visited\_states}

\end{algorithmic}
\end{algorithm}
In Algorithm \ref{alg_update}, we specify the pseudo-code for a function \textit{update}, that given a state $s$, list \textit{actions\_taken}, list \textit{visited\_states}and given an action $a$, appends $a$ to \textit{actions\_taken}. It also appends the list  \textit{actions\_taken} as well as the new resultant state resulting from the action $a(s)$ to the list  \textit{visited\_states}. The list \textit{actions\_taken} is used to track all the actions attempted from the current state to avoid repeating them. The function \textit{update} is called by both functions \textit{intervene} and \textit{make\_consistent}.

\subsubsection{Candidate Path}
Given the CFG $(S_C,S_Q,I,\delta)$ constructed from a run of  algorithm \ref{alg_path} and the return value that we refer to as $r$ such that r is a list (algorithm succeeds). $r'$ is the resultant list obtained from removing all elements containing states $s'\not\in S_C$. We can construct the corresponding candidate path as follows: $r'_i$ represents the $i$ th element of list $r'$. The candidate path is the sequence of states $s_0,...,s_{m-1}$, where $m$ is the length of list $r'$. $s_i$ is the state corresponding to $r'_i$ for $0\leq i<m$.       

\subsection{Proofs}

\begin{theorem}{Soundness Theorem}\label{theorem_soundness}\\
\noindent Given a CFG $\mathbb{X}=(S_C,S_Q,I,\delta)$, constructed from a run of algorithm \ref{alg_path} and a corresponding candidate path $P$, $P$ is a solution path for $\mathbb{X}$.
\begin{proof}
Let $G$ be a goal set for $\mathbb{X}$. By definition \ref{defn:candidate_path}, $P=s_0,...,s_{m}$, where $m\geq0$.
By definition \ref{solution_path}
we must show $P$ has the following properties.
        
        1) $s_0=I $
        
        2) $s_m\in G $
        
        3) $s_j\in S_C\ for\ all\ j\ \in\{0,...,m\}$
        
        4) $s_0,...,s_{m-1} \not\in G$
        
        5) $s_{i+1}\in \delta(s_i)\ for\ i\ \in\{0,...,m-1\}$\\
1) By definition \ref{defn:I}, $i$ is causally consistent and cannot be removed from the candidate path. Hence I must be in the candidate path and is the first state as per line 2 in algorithm \ref{alg_path}. Therefore $s_0$ must be $i$.\\
2) The while loop in algorithm \ref{alg_path} ends if and only if $is\_counterfactual(s,C,Q)$ is True. From theorem \ref{theorem_is_counterfactual}, $is\_counterfactual(s,C,Q)$ is True only for the goal state. Hence $s_m\in G$.\\
3) By  definition \ref{defn:candidate_path} of the candidate path, all states $s_j\in S_C\ for\ all\ j\ \in\{0,...,m\}$.\\
4) By theorem \ref{theorem_all_but_last}, we have proved the claim $s_0,...,s_{m-1} \not\in G$.\\
5) By theorem \ref{theorem_delta}, we have proved the claim $s_{i+1}\in \delta(s_i)\ for\ i\ \in\{0,...,m-1\}$.\\
Hence we proved the candidate path $P$ (definition \ref{defn:candidate_path}) is a solution path (definition \ref{solution_path}).

\end{proof}
\end{theorem}

\begin{theorem}\label{theorem_is_counterfactual}
\noindent Given a CFG $\mathbb{X}=(S_C,S_Q,I,\delta)$, constructed from a run of algorithm \ref{alg_path}, with goal set $G$, and $s\in S_C$; $is\_counterfactual(s,C,Q)$ will be $TRUE$ if and only if $s\in G$.
\begin{proof}

By the definition of the goal set $G$ we have
\begin{equation}
G = \{ s \in S_C| s\not\in S_Q\} \label{theorem_1_1}
\end{equation}
For $is\_counterfactual$ which takes as input the state $s$, the set of causal rules $C$ and the set of decision rules $Q$ (Algorithm \ref{alg_counterfactual}), we see that by from line 1 in algorithm \ref{alg_counterfactual}, it returns $TRUE$ if it satisfied all rules in $C$ and no rules in $Q$.

By the Definition \ref{defn:S_Q}, $s\in S_Q$ \textit{if and only if} it satisfies a rule in $Q$. 
Therefore, $is\_counterfactual(s,C,Q)$ is $TRUE$ if and only if $s\not\in S_Q$ and since $s\in S_C$ and $s\not\in S_Q$, then $s\in G$.

\end{proof}
\end{theorem}

\begin{theorem}\label{theorem_delta}
\noindent Given a CFG $\mathbb{X}=(S_C,S_Q,I,\delta)$, constructed from a run of algorithm \ref{alg_path} and a corresponding candidate path $P=s_0,...,s_{m}$; $s_{i+1}\in \delta(s_i)\ for\ i\ \in\{0,...,m-1\}$

\begin{proof}
This property can be proven by induction on the length of the list \textit{visited\_lists} obtained from Algorithm \ref{alg_path}, \ref{alg_inner_delta}, and \ref{alg_intervene}.\\
\textbf{Base Case}: The list \textit{visited\_lists} from algorithm \ref{alg_path} has length of 1, i.e., [$s_0$]. The property $s_{i+1}\in \delta(s_i)\ for\ i\ \in\{0,...,m-1\}$ is trivially true as there is no $s_{-1}$.\\
\textbf{Inductive Hypotheses}: 
We have a list [$s_0,...,s_{n-1}$] of length $n$ generated from $0$ or more iteration of running the function \textit{intervene} (algorithm \ref{alg_intervene}), and it  satisfies the claim $s_{i+1}\in \delta(s_i)\ for\ i\ \in\{0,...,n-1\}$\\

\textbf{Inductive Step}: If we have a list  [$s_0,...,s_{n-1}$]  of length n and we wish to get element $s_n$ obtained through running another iteration of function \textit{intervene} (algorithm \ref{alg_intervene}). Since [$s_0,...,s_{n-1}$] is  of length n by the inductive hypothesis, it satisfies the property, and it is sufficient to show $s_n\in\delta(s_{n-1})$ where $s_{i+1}\in \delta(s_i)\ for\ i\ \in\{0,...,n-1\}$.\\

The list \textit{visited\_lists} from algorithm \ref{alg_path} has length of $n$. Going from $s_{n-1}$ to $s_n$ involves calling the function \textit{intervene} (Algorithm \ref{alg_intervene}) which in turn calls the function \textit{make\_consistent} (Algorithm \ref{alg_inner_delta}).

Function \textit{make\_consistent} (Algorithm \ref{alg_inner_delta}) takes as input the state $s$, the list of actions taken \textit{actions\_taken}, the list of visited states \textit{visited\_states}, the set of causal rules $C$ and the set of possible actions $A$.
It returns \textit{visited\_states} with the new causally consistent states as the last element. From line 1, if we pass as input a causally consistent state, then function \textit{make\_consistent} does nothing. On the other hand, if we pass a causally inconsistent state, it takes actions to reach a new state. Upon checking if the action taken results in a new state that is causally consistent from the \textit{while} loop in line 1, it returns the new state. 
Hence, we have shown that the moment a causally consistent state is encountered in function \textit{make\_consistent}, it does not add any new state.

Function \textit{intervene} (Algorithm \ref{alg_intervene}) takes as input the list of visited states \textit{visited\_states} which contains the current state as the last element, the set of causal rules $C$ and the set of possible actions $A$. It returns \textit{visited\_states} with the new causally consistent states as the last element. It calls the \textit{make\_consistent} function. For the function \textit{intervene}, in line 1 it obtains the current state (in this case $s_{n-1}$) from the list \textit{visited\_states}. It is seen in line 2 that an action $a$ is taken: 

        1) Case 1: If a causal action is taken, then upon entering the the function \textit{make\_consistent} (Algorithm \ref{alg_inner_delta}), it will not do anything as causal actions by definition result in causally consistent states.         

        2) Case 2: If a direct action is taken, then the new state that may or may not be causally consistent is appended to \textit{visited\_states}. The call to the function \textit{make\_consistent} will append one or more states with only the final state appended being causally consistent.

Hence we have shown that the moment a causally consistent state is appended in function \textit{intervene}, it does not add any new state. This causally consistent state is $s_n$. In both cases $s_n=\sigma(s_{n-1})$ as defined in Definition \ref{defn:impl_cfg} and this $s_n \in \delta(s_{n-1})$.




\end{proof}
\end{theorem}

\begin{theorem}\label{theorem_all_but_last}
\noindent Given a CFG problem $\mathbb{X}=(S_C,S_Q,I,\delta)$, constructed from a run of algorithm \ref{alg_path}, with goal set $G$ and a corresponding candidate path $P=s_0,...,s_{m}$ with $m\geq0$, $s_0,...,s_{m-1} \not\in G$.
\begin{proof}
This property can be proven by induction on the length of the list \textit{visited\_lists} obtained from Algorithm \ref{alg_path}, \ref{alg_inner_delta}, and \ref{alg_intervene}.\\
\textbf{Base Case}: \textit{visited\_lists} has length of 1. 
Therefore the property $P=s_0,...,s_{m}$ with $m\geq0$, $s_0,...,s_{m-1} \not\in G$ is trivially true as state $s_{j}$ for $j<0$ does not exist.\\


\textbf{Inductive Hypotheses}: 
We have a list [$s_0,...,s_{n-1}$] of length $n$ generated from $0$ or more iteration of running the function \textit{intervene} (Algorithm \ref{alg_intervene}), and it  satisfies the claim $s_0,...,s_{n-2} \not\in G$.\\

\textbf{Inductive Step}: Suppose we have a list [$s_0,...,s_{n-1}$] of length n and we wish to append the n+1 th element (state $s_{n}$) by calling the function \textit{intervene}, and we wish to show that that the resultant list satisfies the claim $s_0,...,s_{n-1} \not\in G$. The first n-1 elements ($s_0,...,s_{n-2}$) are not in $G$ as per the inductive hypothesis.

From line 3 in the function \textit{find\_path} (Algorithm \ref{alg_path}), we see that to call the function \textit{intervene} another time, the current state (in this case $s_{n-1})$ \textbf{cannot} be a counterfactual, by theorem \ref{theorem_is_counterfactual}. Hence $s_{n-1}\not\in G$ 

Therefore by induction the claim $s_0,...,s_{n-1}\not \in G$ holds.
\end{proof}
\end{theorem}

\subsection{Experimental Setup}

\subsubsection{Counterfactuals}

Algorithm \ref{alg_counterfactual} \textit{is\_counterfactual}  returns {\tt True} for all states consistent with causal rules $C$ that disagree with the decision rules $Q$. Given our state space $S$, from \textit{is\_counterfactual}, we obtain all states that are realistic counterfactuals. Table \ref{tbl_counterfactual} shows the number of counterfactuals that we obtain using \textit{is\_counterfactual}. 
\begin{table*}[t]
\fontsize{9}{10}\selectfont
\begin{tabular}{cll}
\toprule
Dataset & \# of Features Used & \# of Counterfactuals\\ \midrule
Adult & 6 & 112\\ \cmidrule(lr){1-3}
Cars & 4 & 78 \\ \cmidrule(lr){1-3}
German & 7 & 240 \\ \cmidrule(lr){1-3}
\end{tabular}
\caption{Table showing a Number of Counterfactuals produce by the \textit{is\_counterfactual} function given all possible states.}
\label{tbl_counterfactual}
\end{table*}

\subsubsection{Dataset: Cars }



The Car Evaluation dataset provides information on car purchasing acceptability. We relabelled the Car Evaluation dataset to \textit{`acceptable'} and \textit{`unacceptable'} in order to generate the counterfactuals. We applied the \textit{CoGS} methodology to rules generated by the FOLD-SE algorithm. These rules indicate whether a car is \textit{acceptable} to buy or should be \textit{rejected}, with the undesired outcome being rejection.

For the Car Evaluation dataset, table \ref{tbl_exp}) shows a path from initial \textit{rejection} to changes that make the car \textit{acceptable} for purchase.

We run the FOLD-SE algorithm to produce the following rules: \\
{\tt label(X,`negative') :- persons(X,`2').}\\\\
{\tt label(X,`negative') :- safety(X,`low').}\\\\
{\tt label(X,`negative') :- buying(X,`vhigh'), maint(X,`vhigh').}\\\\
{\tt label(X,`negative') :- not buying(X,`low'), not buying(X,`med'),}\\
{ \tt  \indent\indent\indent\indent\indent\indent\indent\indent maint(X,`vhigh').}\\\\
{\tt label(X,`negative') :- buying(X,`vhigh'), maint(X,`high').}\\

The rules described above indicate if the purchase of a car was rejected . 
 \begin{enumerate}
     \item Accuracy: 93.9\%
     \item Precision: 100\%
     \item Recall: 91.3\%
 \end{enumerate}

2) Features and Feature Values used:
\begin{itemize}
    \item Feature: persons
    \item Feature: safety
    \item Feature: buying
    \item Feature: maint

\end{itemize}

\subsubsection{Dataset: Adult }




The Adult dataset includes demographic information with labels indicating income (‘$=<\$50k/year$’ or ‘$>\$50k/year$’). We applied the \textit{CoGS} methodology to rules generated by the FOLD-SE algorithm on the Adult dataset \cite{adult}. These rules indicate whether someone makes `$=<$\$50k/year'.

Additionally, causal rules are also learnt and verified (for example, if feature \textbf{`Marital Status'} has value  \textit{`Never Married'}, then feature \textbf{`Relationship'} should \textbf(not) have the value  \textit{`husband'} or \textit{`wife'}. We learn the rules to verify these assumptions on cause-effect dependencies.

The goal of \textit{CoGS} is to find a path to a counterfactual instance where a person makes `$>$\$50k/year'. 

For the Adult dataset, Table \ref{tbl_exp} shows a path from making `$=<$\$50k/year' to `$>$\$50k/year'.

We run the FOLD-SE algorithm to produce the following decision making rules: \\
{\tt label(X,`<=50K') :- not marital\_status(X,`Married-civ-spouse'),

\indent\indent\indent\indent\indent\indent\indent\indent capital\_gain(X,N1), N1=<6849.0.}\\\\
{\tt label(X,`<=50K') :- marital\_status(X,`Married-civ-spouse'),

\indent\indent\indent\indent\indent\indent\indent\indent capital\_gain(X,N1), N1=<5013.0,

\indent\indent\indent\indent\indent\indent\indent\indent education\_num(X,N2), N2=<12.0.}\\

 \begin{enumerate}
     \item Accuracy: 84.5\%
     \item Precision: 86.5\%
     \item Recall: 94.6\%
 \end{enumerate}

2) FOLD-SE gives Causal rules for the `marital\_status' feature having  value `never\_married':\\\\{\tt marital\_status(X,`Never-married') :- not relationship(X,`Husband'), 
    
    \indent\indent\indent\indent\indent\indent\indent\indent\indent\indent not relationship(X,`Wife'), age(X,N1), N1=<29.0.} 

 \begin{enumerate}
     \item Accuracy: 86.4\%
     \item Precision: 89.2\%
     \item Recall: 76.4\%
 \end{enumerate}

3) FOLD-SE gives Causal rules for the `marital\_status' feature having  value `Married-civ-spouse':\\\\
{\tt marital\_status(X,`Married-civ-spouse') :- relationship(X,`Husband').} \\\\
{\tt marital\_status(X,`Married-civ-spouse') :- relationship(X,`Wife').} 

 \begin{enumerate}
     \item Accuracy: 99.1\%
     \item Precision: 99.9\%
     \item Recall: 98.2\%
 \end{enumerate}

4) For values of the feature `marital\_status' that are not `Married-civ-spouse'  or `never\_married' which we shall call `neither', a user defined rule is used:\\\\
{\tt marital\_status(X,neither) :- not relationship(X,`Husband'),

\indent\indent\indent\indent\indent\indent\indent\indent not relationship(X,`Wife').}

5) FOLD-SE gives Causal rules for the `relationship' feature having  value `husband':\\\\
{\tt relationship(X,`Husband') :- sex(X,`Male') ,

\indent\indent\indent\indent\indent\indent\indent\indent age(X,N1), not(N1=<27.0).} 

 \begin{enumerate}
     \item Accuracy: 82.3\%
     \item Precision: 71.3\%
     \item Recall: 93.2\%
 \end{enumerate}
 
5) For the  `relationship' feature value of `wife', a user defined rule is used:\\\\
{\tt relationship(X,`Wife') :- sex(X,`Female').} 
    
6)Features Used in Generating the counterfactual path:
\begin{itemize}
    \item Feature: marital\_status
    \item Feature: relationship
    \item Feature: sex

    \item capital\_gain
    \item education\_num
    \item age
    
\end{itemize}

\subsubsection{Dataset: German }
We run the FOLD-SE algorithm to produce the following decision making rules: \\\\
{\tt label(X,`good'):-checking\_account\_status(X,`no\_checking\_account')}\\\\
{\tt label(X,`good'):-not checking\_account\_status(X,`no\_checking\_account'),}\\
{\tt \indent\indent\indent\indent\indent\indent  not credit\_history(X,`all\_dues\_atbank\_cleared'),} \\
{\tt \indent\indent\indent \indent\indent\indent duration\_months(X,N1), N1=<21.0,}\\
{\tt \indent\indent\indent \indent\indent\indent credit\_amount(X,N2), not(N2=<428.0),}\\
{\tt \indent\indent\indent \indent\indent\indent not ab1(X,`True').}\\\\
{\tt ab1(X,`True'):-property(X,`car or other'), \\
\indent\indent\indent \indent\indent\indent credit\_amount(X,N2), N2=<1345.0.}
    
 \begin{enumerate}
     \item Accuracy: 77\%
     \item Precision: 83\%
     \item Recall: 84.2\%
 \end{enumerate}

2) FOLD-SE gives Causal rules for the `present\_employment\_since' feature having  value `employed' where employed is the placeholder for all feature values that are \textbf{not} equal to the feature value `unemployed':

    {\tt present\_employment\_since(X,`employed') :- \\
    \indent\indent\indent\indent\indent\indent not job(X,`unemployed/unskilled-non\_resident').

 \begin{enumerate}
     \item Accuracy: 95\%
     \item Precision: 96.4\%
     \item Recall: 98.4\%
 \end{enumerate}

3) For values of the feature `present\_employment\_since' that are `unemployed', a user defined rule is used

    {\tt present\_employment\_since(X,`unemployed') :- \\
    \indent\indent\indent\indent\indent\indent job(X,`unemployed/unskilled-non\_resident').

6)Features Used in Generating the counterfactual path:
\begin{itemize}
    \item checking\_account\_status
    
    \item credit\_history
    
    \item property

    \item duration\_months
    
    \item credit\_amount
    
    \item present\_employment\_since

    \item job 
\end{itemize}

\end{document}